\documentclass[journal]{IEEEtran}
\ifCLASSINFOpdf
\else
\fi

\usepackage{lineno,hyperref}
\usepackage{graphicx}
\usepackage{subfig}
\usepackage{epsfig}
\usepackage{multicol}
\usepackage{float}
\usepackage{bm}
\usepackage{amsmath}
\usepackage{threeparttable}
\usepackage[table,xcdraw]{xcolor}
\usepackage{multirow}
\usepackage{stfloats}
\usepackage{amssymb}
\usepackage{color}
\begin{document}
%
\title{Temporal Graph Modeling for Skeleton-based Action Recognition}
%
%
%

\author{Jianan~Li, Xuemei~Xie, Zhifu~Zhao, Yuhan~Cao, Qingzhe~Pan and Guangming~Shi~\IEEEmembership{Fellow,~IEEE}
	\thanks{
		Jianan~Li, Xuemei~Xie, Zhifu~Zhao, Yuhan~Cao, Qingzhe~Pan, Guangming~Shi are with School of Artificial Intelligence, Xidian University, Xi'an, China. E-mail: jiananli.xd@gmail.com; xmxie@mail.xidian.edu.cn.
	}}
\maketitle

\begin{abstract}
Graph Convolutional Networks (GCNs), which model skeleton data as graphs, have obtained remarkable performance for skeleton-based action recognition.
Particularly, the temporal dynamic of skeleton sequence conveys significant information in the recognition task. For temporal dynamic modeling, GCN-based methods only stack multi-layer 1D local convolutions to extract temporal relations between adjacent time steps. With the repeat of a lot of local convolutions, the key temporal information with non-adjacent temporal distance may be ignored due to the information dilution.
Therefore, these methods still remain unclear how to fully explore temporal dynamic of skeleton sequence.
In this paper, we propose a Temporal Enhanced Graph Convolutional Network (TE-GCN) to tackle this limitation. The proposed TE-GCN constructs temporal relation graph to capture complex temporal dynamic. Specifically, the constructed temporal relation graph explicitly builds connections between semantically related temporal features to model temporal relations between both adjacent and non-adjacent time steps. Meanwhile, to further explore the sufficient temporal dynamic, multi-head mechanism is designed to investigate multi-kinds of temporal relations. Extensive experiments are performed on two widely used large-scale datasets, NTU-60 RGB+D and NTU-120 RGB+D. And experimental results show that the proposed model achieves the state-of-the-art performance by making contribution to temporal modeling for action recognition.
\end{abstract}


%
\IEEEpeerreviewmaketitle

\section{Introduction}\label{Introduction}
Action recognition is a highly important direction in computer vision, which has extensive applications from intelligent surveillance and virtual reality to human robot interaction. According to the type of input data, the research on action recognition can be categorized into skeleton-based and RGB-based approaches.
In particular, compared with RGB-based approaches, skeleton-based approaches \cite{zhang2017view, liu2017skeleton,yan2018spatial} become an attractive research direction due to their robustness of variations in background, illumination and viewpoint.
Recently, with the development of cost-effective depth cameras \cite{shotton2011real} and pose estimation algorithms \cite{newell2016stacked}, skeleton-based action recognition becomes more and more popular from scholars.

In the task of skeleton-based action recognition, we are given a time series of human joint coordinates (skeleton sequence) and expected to model the motion pattern thereon. Early deep-learning based approaches \cite{hussein2013human,vemulapalli2014human,veeriah2015differential} usually rearrange the skeleton sequences into 2D gridded data and then directly feed them into Recurrent Neural Networks (RNNs) or Convolution Neural Networks (CNNs).
Nevertheless, human skeleton is naturally structured as a graph instead of gridded data, which makes it difficult for RNNs and CNNs based approaches to fully exploit the structure information of human skeleton.
In recent years, Graph Convolutional Networks (GCNs)~\cite{yan2018spatial, shi2019two, li2019spatio} process skeleton data in a flexible way and introduce various incremental modules to enhance the expressive power of skeleton structure.
\begin{figure}
	\centering
	\includegraphics[width=0.9\linewidth]{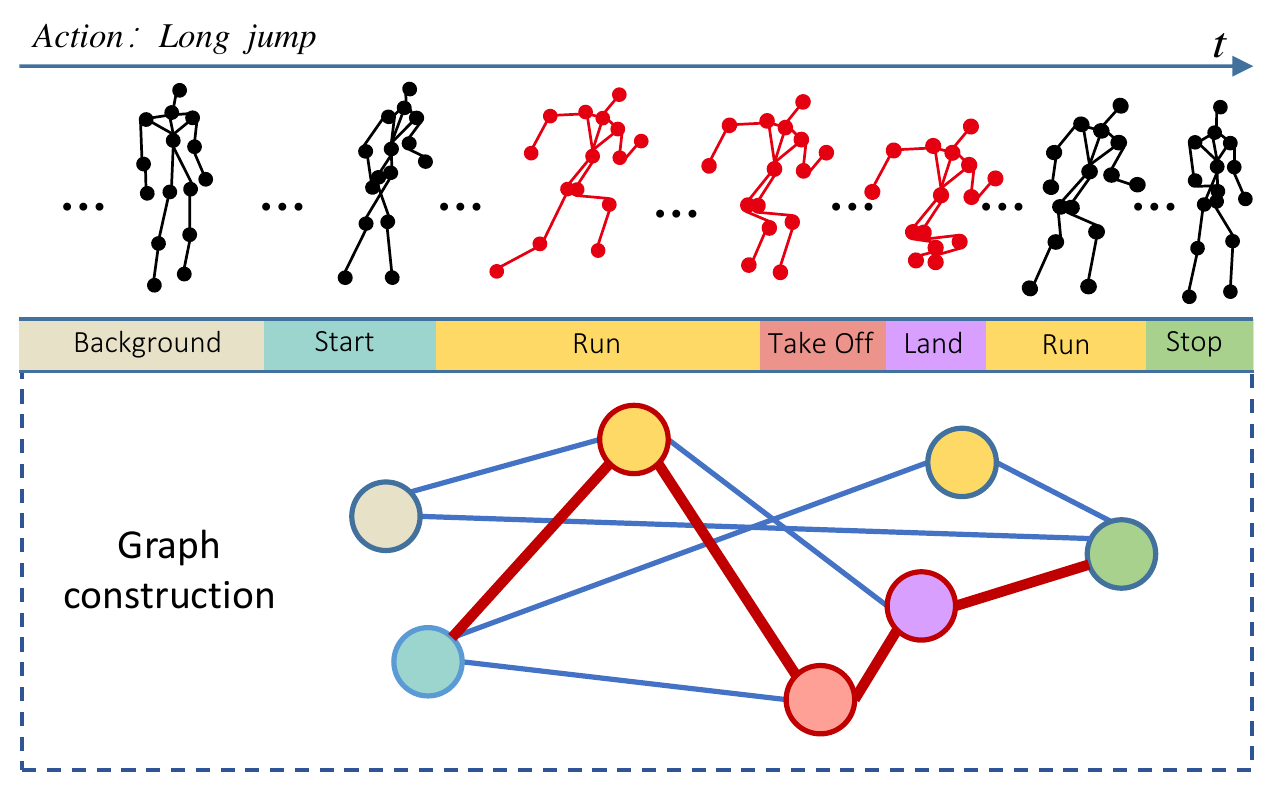}
	\caption{The example of the proposed temporal relation graph to represent temporal relations. Considering the example of ``\textbf{\emph{Long jump}}'', it contains the sub-actions like ``Start'', ``Run'', ``Take off'' and ``Land''. The constructed temporal relation graph directly construct correlation among key sub-actions ``Run'', ``Take off'' and ``Land'' and further decide the action ``Long jump''. Thus, the long-range dependencies of non-adjacent time steps can be captured, which can promote the action recognition.}
	\label{fig:example}
\end{figure}
Though these methods have obviously improvements in modeling sufficient relations within skeleton structure, it remains unclear how to model complex temporal dynamic of skeleton sequence for action recognition.
Specifically, GCN-based methods deploy graph convolution to extract spatial relation at individual frames and then use 1D local convolution to model temporal dynamic, which is analogous to factorized 3D convolution~\cite{qiu2017learning, tran2018closer}.
The 1D local convolution only explicitly models correlation between adjacent time steps in skeleton sequence.
To capture temporal dynamic of whole skeleton sequence, it is common to stack multi-layer local convolutions.
However, with the repeat of a lot of local convolutions, the key temporal information with non-adjacent temporal distance may be ignored due to the information dilution. Even the action-irrelevant temporal information in local distant may interfere with the feature aggregation for temporal modeling.
Therefore, GCN-based methods fail to fully explore the temporal dynamic of skeleton sequence.

{As we know, an ordinary action consists of different time steps.
The correlations among only several key time steps with adjacent or non-adjacent temporal distance may contribute to the action recognition.
As the example shown in Fig.~\ref{fig:example}, the action ``\textbf{\emph{Long jump}}'' contains the time steps like ``Start'', ``Run'', ``Take off'' and ``Land''. And the correlations among key time steps ``Run'', ``Take off'' and ``Land'' decide the action ``Long jump''.
Therefore, in temporal modeling, the correlations between key time steps, especially in non-adjacent temporal distance, should be explicitly exploited in a flexible method.}

In this paper, we propose Temporal Enhanced Graph Convolutional Network (TE-GCN) to construct the graph structure on temporal dimension for skeleton-based action recognition. Specifically, we construct the temporal relation graph to directly capture the temporal dynamic between both adjacent and non-adjacent time steps. In each temporal relation graph, the node represents the temporal feature and the edge represents the temporal relation. For edge construction, two transform functions, Feature Calculated and Feature Learned, are devised to calculate the correlations between temporal features.
Additionally, there may be multiple relationships between nodes and their temporal neighbors, which shows different importance in learning discriminative features.
To investigate multi-kinds of temporal relations between different time steps, the multi-head temporal relation graph is designed with multi-head mechanism. And then, the multi-head temporal enhanced graph convolution aggregates these multi-kinds relations and enables to capture sufficient temporal relations of non-adjacent time steps for action recognition.
Besides, we also employ multi-stream TE-GCN to explore multi-modalities (the spatial information of the joint, bone as well as their corresponding motion information) for further performance improvement.

To verify the effectiveness of the proposed method, exhaustive experiments are performed on two prevalence datasets: NTU-60 RGB+D and NTU-120 RGB+D. Experimental results show that the proposed final model achieves the state-of-the-art performance for skeleton-based action recognition.
Our major contributions are summarized as follows:
\begin{enumerate}
  \item For modeling complex temporal dynamic of skeleton sequence, we propose a novel temporal graph convolution to explicitly capture correlations between both adjacent and non-adjacent time steps.
  \item We construct a multi-head TE-GCN to explore multi-kinds of temporal relations between different time steps, which is important for action recognition.
  \item Extensive experiments are performed on two prevalence datasets (NTU-60 and NTU-120 RGB+D datasets) to verify the effectiveness of the proposed method.
\end{enumerate}

\section{Related Work}\label{RelatedWork}
\subsection{Skeleton-based action recognition}

Due to the robustness to variations in background, illumination and viewpoint, skeleton-based action recognition becomes an attractive research direction in computer vision. Besides, in recent years, the development of cost-effective depth cameras and pose estimation algorithms makes it easier to obtain human skeleton information. Therefore, skeleton-based action recognition has recently attracted substantial attention from scholars.

Earlier approaches \cite{hussein2013human,vemulapalli2014human,veeriah2015differential} investigate the dynamic of human skeleton sequence by designing hand-crafted features.
For example, covariance matrix of joint locations over time is used in \cite{hussein2013human} as a discriminative descriptor for a skeleton sequence.
However, the hand-crafted features are not flexible enough to model discriminative feature of the action.
In contrast, the deep-learning models can automatically learn suitable features in an end-to-end manner. With the development of the deep-learning methods, recent approaches \cite{zhang2017view,liu2017skeleton, yan2018spatial, shi2019two, li2019spatio} achieve impressive performance for skeleton-based action recognition. These methods can be summarized as three flows: Convolution Neural Networks (CNNs), Recurrent Neural Network (RNNs) and Graph Convolutional Networks (GCNs) based methods.

The CNNs and RNNs are designed to process the grid-shaped data. However, skeleton data is in the form of graph.
To benefit from the strong representation ability of RNNs or CNNs, early deep-learning based methods \cite{zhang2017view, liu2017skeleton, li2018independently} rearrange the skeleton data into 2D gridded data and then directly feed them into RNNs and CNNs.
These methods, as mentioned in \cite{monti2017geometric}, cannot fully exploit the structure information of skeleton data. Recently, GCN-based methods \cite{yan2018spatial, shi2019two, li2019spatio, li2019actional, zhang2019graph} process skeleton data in a flexible way to explore the relations between skeleton joints. Yan et al.~\cite{yan2018spatial} are the first to use GCN for skeleton-based action recognition. ST-GCN \cite{yan2018spatial} uses spatial graph convolution along with temporal convolution for motion representation. To further explore the diversity of skeleton data for action recognition task, Two-stream adaptive GCN \cite{shi2019two} introduces adaptive graph with self-attention and a freely learned graph adjacent matrix mask. Similarly, AS-GCN \cite{li2019actional} also uses adjacency powering for multi-scale modeling. Besides, Zhang et al.~\cite{zhang2019graph} investigate body bones from skeleton data and develop convolutions over graph edges corresponding to the bones in human skeleton.
However, these methods mentioned above fail to consider how to explore the temporal relation of skeleton sequence.

\subsection{Graph Convolution Networks}

Convolution Neural Networks (CNNs) have gained great success in processing grid-shaped data, such as image, video and audio. However, it is not straight forward to use CNNs to process graph data. In the past decades, how to operate on graph data has been extensively explored. Currently, the most effective way is to use Graph Convolutional Networks (GCNs). GCNs generalize CNNs to process data in graphs with arbitrary structures and shapes. Therefore, it possesses the strong representation ability to handle graph data. According to the definition of the graph convolution, the GCNs mainly fall into two streams: spatial perspective and spectral perspective. The spatial perspective methods directly perform convolution on the graph node and its neighbors. Niepert et al.~\cite{niepert2016learning} sample the neighborhoods for each of the nodes based on their distances in the graph. To further improve the representation of GCN, Graph Attention Network (GAT)~\cite{velivckovic2017graph} introduces the attention mechanisms to GCN. They select information which is relatively critical from all inputs. Inspired by this, Sankar et al.~\cite{sankar2018dynamic} employ self-attention on both spatial and temporal dimensions and get superior results.

By contrast, the spectral perspective methods generate the graph representation in graph Fourier transform domain based on eigen-decomposition. For example, Defferrard et al.~\cite{defferrard2016convolutional} design fast localized convolutional filters by using recurrent Chebyshev polynomials. Kipf et al.~\cite{kipf2016semi} further simplify this approach using the first-order approximation of the spectral graph convolutions. However, these methods are limited by the computational efficiency due to the requirement of eigen-decomposition. Our work follows the spatial perspective methods.

\subsection{Temporal relation modeling in action recognition}
Temporal relation modeling in videos is critical for action recognition.
Particularly, the complex temporal structure of action instance increases the challenge of action recognition.
In recent years, researchers have started to explore this direction. Non-local Neural Networks~\cite{wang2018non} present non-local operations as a generic family of building blocks for capturing long-range temporal dependencies. Meanwhile, to capture high-order interactions between non-adjacent time steps, Non-local Recurrent Neural Memory~\cite{fu2019non} performs non-local operations to learn full-order interactions within a sliding temporal block and models global interactions between blocks in a gated recurrent manner. In addition, Zhou et al.~\cite{zhou2018temporal} propose Temporal Relation Network (TRN) to give CNNs a remarkable capability to discover temporal relations in video. Nevertheless, limited to the local operation of CNNs or RNNs, these methods only implicitly model the long-range temporal relation and fail to fully explore the complex temporal structure of the action. Our work aims to exploit the graph convolutional network over temporal dimension to fully explore the temporal structure of time steps and adaptively learn various temporal relations in videos with a supervised learning setting.
\section{Preliminary}\label{section3}

In this section, we will briefly recapitulate the basic graph convolutional network necessary for the rest of the paper.

\subsection{Notations}\label{subsection31}

The skeleton data represents human actions as a sequence of skeleton frames. For each frame, the skeleton of each person can be denoted as a graph $G = (V,E)$, where $V = \{ {v_i}|i = 1, \ldots ,J\} $ is the set of $J$ human joints and $E = \{ {e_{i,j}}|{v_{i,}}{v_j}\}$ is the set of edges representing the human bones. Mathematically, the structure of skeleton graph can be represented by the adjacent matrix $A \in {\mathbb{R}^{J \times J}}$, with ${A_{ij}} \in \left\{ {1,0} \right\}$ indicating whether an edge exists between joint ${v_i}$ and ${v_j}$. And the sequence of skeleton frames can be represented as a feature tensor $X \in {\mathbb{R} ^{C \times T \times J}}$, where $C$ represents the coordinate dimension (e.g. 3 for 3D skeleton), $T$ is the number of skeleton frames in the sequence, and $J$ denotes the total number of human joints.

\subsection{Graph Convolutional Networks}

Graph Convolutional Networks (GCNs) are the effective frameworks for learning representation of graph structured data.
Given the notations defined above, GCNs perform multiple graph convolutional layers on skeleton sequence to extract the high-level features. Each graph convolutional layer is typically composed of spatial graph convolution block (SG-block) and temporal convolution block (TC-block). Mathematically, the operation of SG-block can be formulated as:
\begin{equation}\label{equation1}
{f_{out}^s = \sum\limits_k^{{K_s}} {{W_k}{f_{in}}(\Lambda _k^{ - \frac{1}{2}}{{{\rm{\tilde A}}}_{\rm{k}}}\Lambda _k^{ - \frac{1}{2}} \odot {M_k})}},
\end{equation}
{\noindent where ${f_{in}}$ denote input feature. And ${f_{out}^s}$ represents output feature of SG-block. ${K_s}$ denotes the number of partition subsets. With the mapping strategy according to ST-GCN \cite{yan2018spatial}, ${K_s}$ is set to 3. ${\rm{\tilde A}} = {\rm{A + I}}$ is the adjacent matrix with added self-loops for keeping identity mapping. $\Lambda$ represents the degree matrix of ${\rm{\tilde A}}$. For each adjacency matrix, we accompany it with a learnable matrix ${M_k} \in {\mathbb{R} ^{J \times J}}$, which can adjust the importance of edges.  The notation $\odot$ is the Hadamard Product. ${W_k}$ denotes the convolution kernel with the size of $1 \times 1$. Note that the number of joints retains unchanged with the convolution of graph.}

After the feature aggregation of SG-block, the TC-block performs normal ${K_t} \times 1$ convolution to extract temporal information of the feature map  ${f^s_{out}}$, where ${K_t}$ is the temporal kernel size. Both SG-block and TC-block are followed by a BatchNorm layer and a ReLU layer. To learn both spatial and temporal features, a GCN is generally constructed by stacking SG-blocks and TC-blocks alternately.
\section{Temporal Enhanced GCN}\label{subsection4}
To capture the complex temporal relations in skeleton sequence and directly build the correlations between non-adjacent time steps, we propose Temporal Enhanced GCN (TE-GCN) for skeleton-based action recognition. In this section, we first give the motivation of the proposed method. And then, the detail about the construction of temporal relation graph is described. Subsequently, the operation of temporal enhanced graph convolution are introduced. Finally, we describe the architecture of the proposed multi-stream TE-GCN.
\subsection{Motivation}\label{subsection41}
With the above discussion, existing GCN-based methods model temporal dynamic information only by repeatedly stacking normal 1D temporal convolutions in deep networks. Generally, the discriminative time steps may occur sparsely in few frames within consecutive skeleton frames of a video. The 1D local convolution only explicitly models correlation between adjacent time steps in skeleton sequence. By repeating a lot of local convolutions between adjacent time steps, the key temporal information may be ignored due to long-range temporal aggregation.
Nevertheless, within an ordinary action, the discriminative features that contribute to action recognition are commonly extracted from the semantically related time steps with non-adjacent distance. For example, the action ``Long jump'' is decided by the key time steps ``Run'', ``Take off'' and ``Land''.
Therefore, it is important to handle the correlations between key time steps for action recognition.

Existing GCN-based methods only consider the spatial relations of skeleton joints in individual frames regardless of the temporal structure in skeleton sequence.
For recognition, the action in video is mainly characterized by the temporal dynamics rather than the static appearance.
To explicitly explore the correlations between non-adjacent time steps in a flexible way, we broadcast the idea of constructing graph from spatial dimension to temporal dimension.
Inspired by the method that constructs spatial relations in each skeleton frame (described in Section~\ref{section3}), we construct a graph ${G_t} = (\chi ,{E_t})$ over temporal sequence, where $\chi $ is the node set, and each node represents a temporal semantic instance. ${E_t}$ is the edge set representing pairwise correlations between temporal semantic instances.
For the edges in temporal relation graph, we explore two schemes to investigate the semantic information of temporal correlations.
The details about constructing the edges of graph will be elaborately  described in the following Section~\ref{subsection42}.

\subsection{Temporal relation graph construction}\label{subsection42}
To explicitly form the correlations between non-adjacent time steps, we propose to construct temporal relation graph along the temporal dimension of skeleton sequence.
For different actions, the obtained temporal relations should be different. Even for different samples under the same action, the obtained temporal relations should be various. The reason is that the action performed by different subjects under the same class may vary a lot. Thus, we propose to construct temporal relation graph whose topology is adaptive for different samples.

Specifically, we denote the input feature as $X \in {\mathbb{R}^{C \times T \times J}}$ (described in Section \ref{subsection31}).
In order to capture temporal relations between time steps in skeleton sequence, it is expected to discover temporal semantic information of skeleton sequence beyond individual frames.
Therefore, to obtain temporal semantic information, we first feed the input feature into regular graph convolution layer to obtain the semantic feature map ${f^l}$ in $l$-th layer.
The feature map ${f^l}$ is in the shape of ${C^l} \times {T^l} \times J$, where ${C^l}$ denotes the channel of feature map, ${T^l}$ denotes the length of temporal dimension and $J$ denotes the total number of human joints of each person.
$f_t^l\in {\mathbb{R}^{C^l \times J}}$ is the $t$-th temporal semantic instance along the temporal dimension. It is denoted as temporal feature. Thus $\chi  = \left\{ {f_t^l} \right\}_{t = 1}^{T^l}$ is the node set of the temporal relation graph. Note that, in order to make temporal feature obtain sufficient semantic information, at least one layer of graph convolution ($l \ge 1$) should be performed to transform input feature into high-level features.
\begin{figure}
	\centering
	\includegraphics[width=1.05\linewidth]{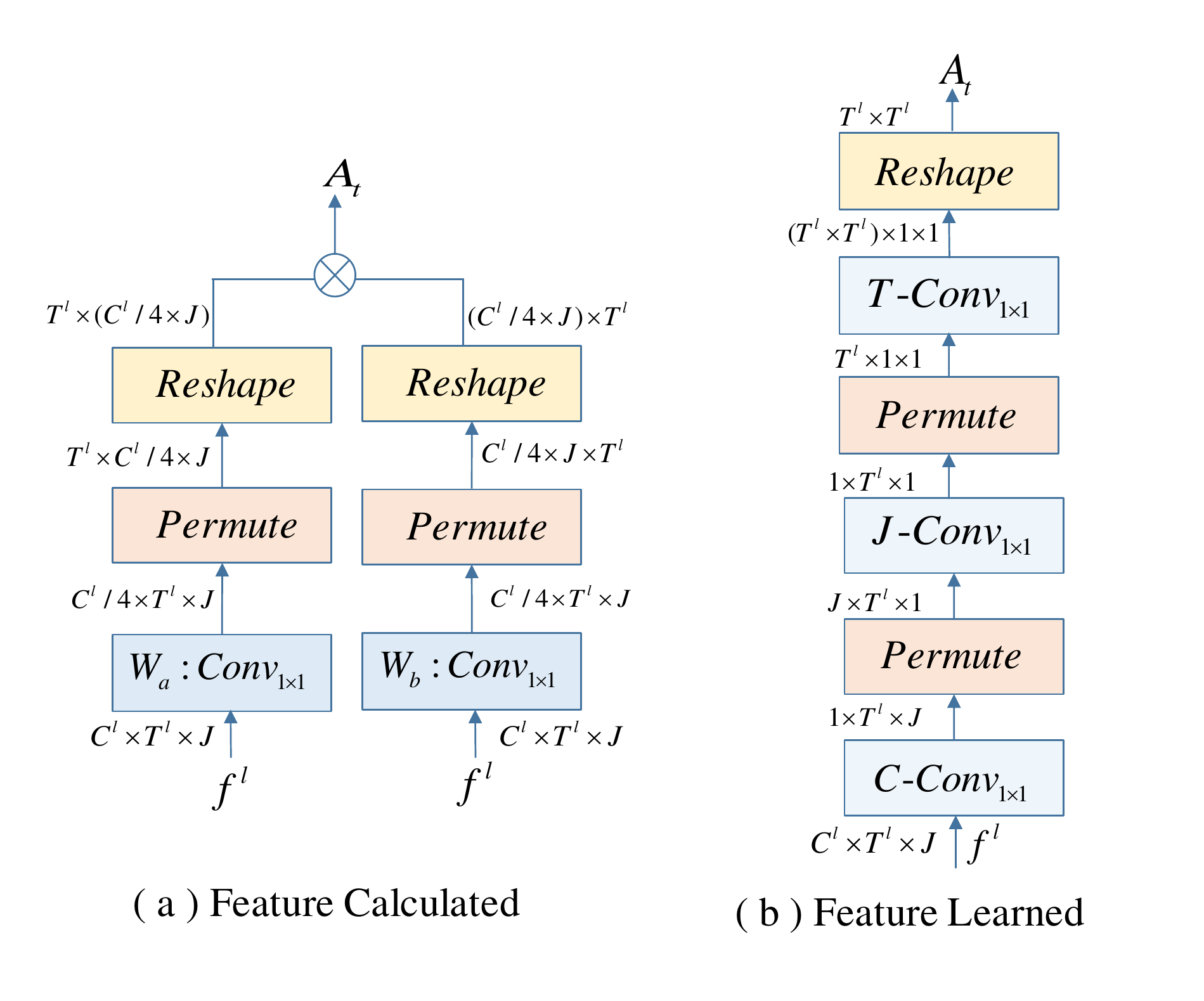}
	\caption{An illustration of the implement architectures about the two different relevant functions. Plot(a) shows the architecture of Feature Calculated. Plot(b) shows the architecture of Feature Learned. The input feature $f^l$ of skeleton sequence  is shown in the tensor format. And the output is the temporal adjacent matrix $ A_t$.}
	\label{fig:feature_correlation}
\end{figure}

After determining the node set of temporal relation graph, we need to consider how to build edges for temporal relation graph.
The edges determine whether there is a connection between temporal features and how strong the connection is.
Considering that the temporal features with related semantic information should be directly aggregated for feature representation, we build the temporal edges according to the defined pairwise correlations between temporal features.

To determine the temporal correlations, two relevant functions, Feature Calculated and Feature Learned, are designed to
measure the relevance between temporal features. The implementation architectures of these two relevant functions are shown in Fig.~\ref{fig:feature_correlation}. More specifically, the detailed descriptions about Feature Calculated and Feature Learned are as follows:
\begin{figure*}[t]
	\centering
	\includegraphics[width=0.75\linewidth]{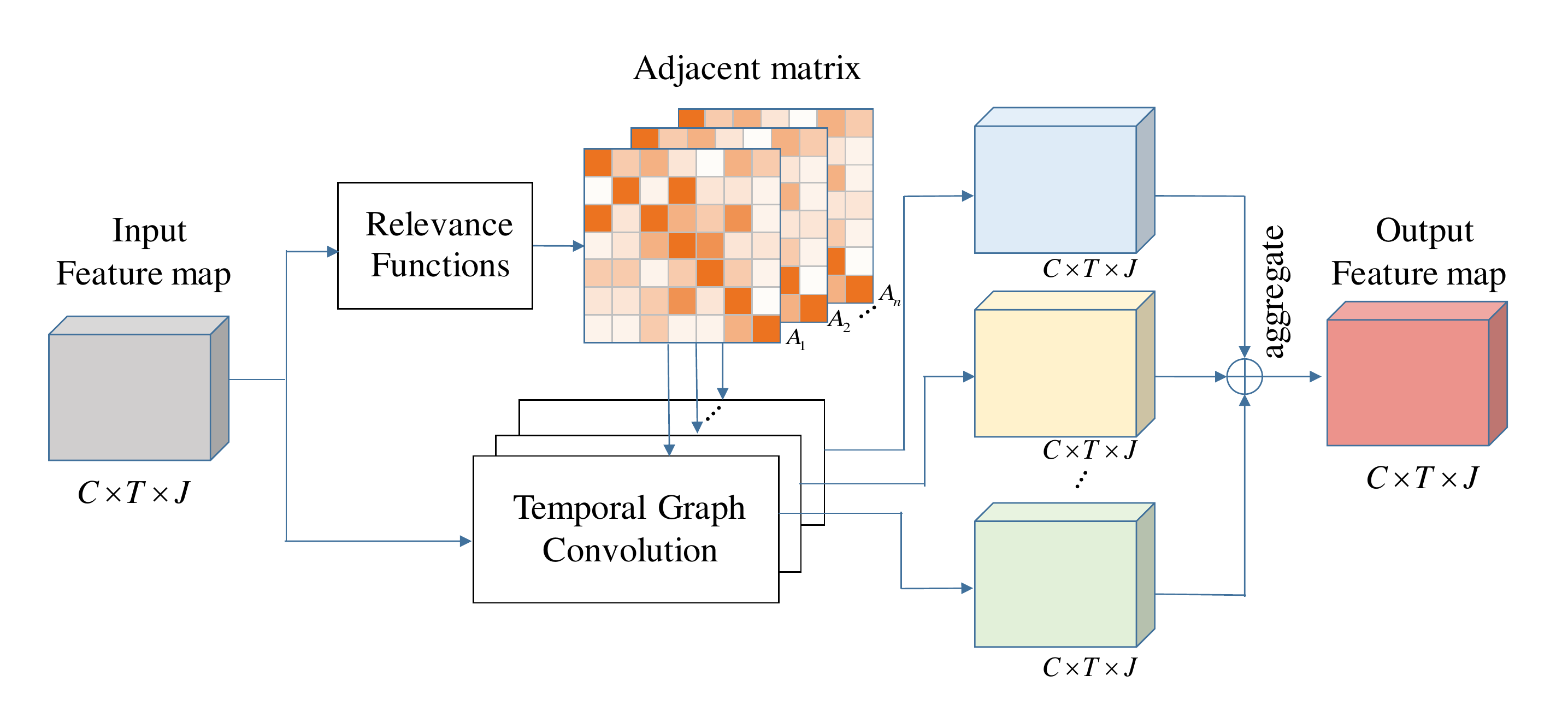}
	\caption{The illustration of the proposed multi-head temporal graph convolution. After obtaining skeleton feature by graph convolution, we first utilize designed relevance function to obtain multi-head adjacent matrices. Subsequently, the temporal graph convolution is performed on individual temporal relation graphs using corresponding temporal adjacent matrix to extract multi-kinds of temporal relations. Finally, the obtained multi-kinds of features are aggregated by element-wise sum to get temporal enhanced features.}
	\label{fig:architecture}
\end{figure*}
\noindent{(1)	Feature Calculated}

The edge connections in temporal relation graph as well as their weights should be different since the correlations are different for pairs of different semantic features. In the Feature Calculated, the different correlations are explicitly calculated by the similarity between temporal features.
Specifically, given the input skeleton sequence with $T$ frames and $J$ joints, the obtained feature map $f^l$ is in the shape of ${C^l} \times {T^l} \times J$.
Then, we utilize $W_a$ and $W_b$ to convert the input feature into the transformed spaces for further calculating the correlation. The $W_a$ and $W_b$ are the trainable transform functions, which are implemented by the $1\times1$ convolutions.
The output features are in the shape of ${C^l/4}\times {T^l} \times J$. Subsequently, to split out each temporal feature, we permute and reshape the feature maps into ${T^l} \times ({C^l/4} \times J)$ and  $({C^l/4} \times J) \times {T^l} $, respectively.
Meanwhile, the corresponding feature vector along temporal dimension $f_{{t}}^l \in {\mathbb{R}^{({C^l/4} \times J)}}$ is temporal feature.
Finally, the correlation operation is performed on each pair of temporal features to measure the correlation:
\begin{equation}\label{equation2}
{r_{ij}}{\rm{ = }}Cor(f_i^l,f_j^l) = {(f_i^l)^{\rm T}}f_j^l,\begin{array}{*{20}{c}}
{}
\end{array}i,j \in [1,{T^l}],
\end{equation}
where $f_i^l$ and $f_j^l$ denote the pairwise temporal features. The obtained correlation scores of each pairwise temporal features are considered as the elements of temporal adjacent matrix $A_t=(r_{ij})_{{T^l}\times{T^l}}$.
It represents the structure information of temporal relation graph.
An implementation example of Feature Calculated is shown in Fig.~\ref{fig:feature_correlation}(a). The Feature Calculated is an explicit manner to build the connection between correlated temporal features.

\noindent{(2)	Feature Learned}

The Feature Learned operation aims to obtain the temporal graph totally learned from training data.
Different from the Feature Calculated (described above), the Feature Learned implicitly obtains the temporal relation in a more flexible way. The architecture of Feature Learned is illustrated in Fig.~\ref{fig:feature_correlation}(b).
Specifically, given the input feature $f^l$ with shape ${C^l} \times {T^l} \times J$, we first squeeze the feature in both channel and spatial dimensions by C-conv and J-conv to obtain the corresponding global representation. Then, the temporal dimension is permuted instead of the channel dimension, and T-conv is performed to obtain global representation of temporal dimension. The C-conv, J-conv and T-conv are all performed by $1 \times 1$ convolutions for feature transform.
Finally, the obtained feature is in the shape of ${T^l} \times {T^l}$, which can be directly denoted as temporal adjacent matrix $A_t$. The temporal adjacent matrix $A_t=(r_{ij})_{{T^l}\times{T^l}}$ is able to represent the edge connections of temporal relation graph.
Each element $r_{ij}$ indicates whether there exists connection between different temporal features as well as the strength of connection.
Note that there are no extra constrains in relevance operation, which means the temporal relation graph is totally learned from the training data.
With this data-driven method, the generated temporal relation graph can be directly targeted to the final recognition task.

By using Feature Calculated and Feature Learned to calculate the correlations between different temporal features, we are able to obtain different temporal adjacent matrices.
Then, normalization is performed across each row of the temporal adjacent matrix so that the sum of all the relevance score corresponding to one temporal feature is 1.
This makes it easy to perform the comparison across different temporal features. Mathematically, we adopt the $softmax$ function for normalization as:
\begin{equation}\label{equation3}
{{a_{ij}} = softmax({r_{ij}}) = \frac{{\exp ({r_{ij}})}}{{\sum {_{k = 1}^{{T^l}}\exp ({r_{ik}})} }}} .
\end{equation}

With temporal adjacent matrix, the generated temporal graph is able to represent the semantic structure of non-adjacent temporal sequence for the specific action. Thus, the aggregated feature can be extracted without the interference of action-irrelevant information.

It should be noted that every skeleton sequence of an action contains multiple types of temporal semantic relations,  which shows different importance in learning discriminative temporal features. However, temporal features extracted from one specific temporal relation graph only reflect temporal semantic relations from one aspect. In order to capture sufficient temporal semantic relations of time steps, we design a multi-head mechanism for temporal relation graph.

\noindent {\textbf{Multi-head temporal relation graph.}} To explore the sufficient temporal relations, we construct multi-head temporal relation graph inspired by the graph attention network~\cite{velivckovic2017graph}. Specifically, $N$ independent relevant functions are performed to obtain $N$ different temporal adjacent matrices. This operation can be formulated as:
\begin{equation}\label{equation4}
{a_{ij}^n = softmax(r_{ij}^n) = softmax[Co{r_n}(f_i^l,f_j^l)]} ,
\end{equation}
{\noindent where $n \in \left\{ {1, 2, \cdots ,N} \right\}$ denotes the index of head. The obtained different correlation coefficients could reflect various correlations between temporal instances. All the temporal adjacent matrices can be denoted as $ \left\{ {A_t^1,A_t^2,\cdots, A_t^n,\cdots,A_t^N} \right\}$, where $A_t^n = {({a_{ij}^n} )_{{T^l} \times {T^l}}}$. Therefore, the multi-head temporal graphs represented by temporal adjacent matrices are able to investigate multiple relations between different temporal instances.}

\subsection{Temporal enhanced graph convolution}
Once getting the multi-head temporal relation graphs, we propose the multi-head temporal graph convolution to explore how to extract sufficient temporal features from these temporal relation graphs.
The overall architecture of the multi-head temporal graph convolution is shown in Fig.~\ref{fig:architecture}, which is the basic building block of the Temporal Enhanced GCN framework. Specifically, the temporal graph convolution is firstly performed on individual temporal graphs for feature extraction. Then, the obtained multi-kinds of features are aggregated by element-wise sum to get sufficient temporal features.

\begin{figure*}
	\centering
	\includegraphics[width=0.8\linewidth]{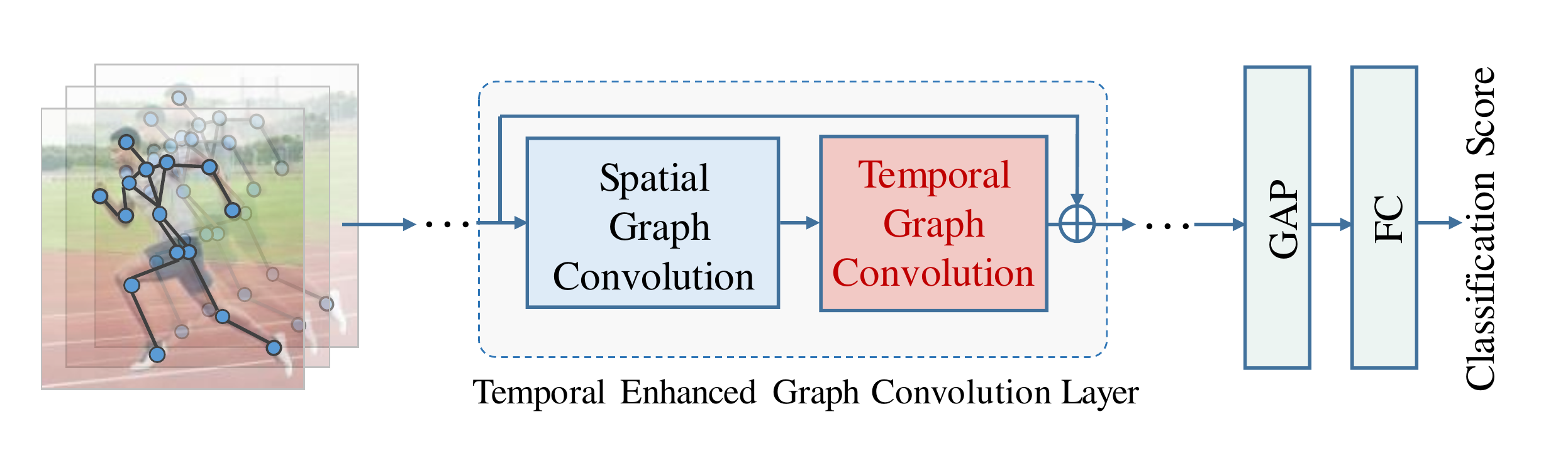}
	\caption{Architecture overview of Temporal Enhanced Graph Convolutional Network. Each temporal Enhanced Graph Convolutional layer is the combination of regular spatial graph convolution and temporal graph convolution in a residual manner.}
	\label{fig:Architecture}
\end{figure*}
For temporal graph convolution, we adopt the output feature of SG-block as the input and then perform graph convolution along temporal dimension to get the new sequence representation.
In the new sequence representation, the structure of temporal graph is maintained, but the meaning of each node is changed. Each node aggregates the feature from its neighbor set according to the designed temporal graph.
Therefore, the temporal graph convolution is able to directly aggregate features from adjacent and non-adjacent temporal instances and capture the structure information in temporal dimension.
Besides, multi-head temporal graphs represent multiple correlations between temporal features. These multiple relationships will show different importance in learning discriminative temporal features. In order to learn comprehensive temporal features, the obtained features in terms of individual temporal graphs are fused by element-wise sum. The multi-head temporal graph convolution can be formulated as:
\begin{equation}\label{equation5}
{{f_{out}} = \sum\limits_{n=1}^N {{\rm{A}}_t^n{f^s_{out}}} {W_t^n}},
\end{equation}
{\noindent where $f^s_{out}$ denotes the output feature of SG-block. ${\rm{A}}_t^n$ represents one of the multi-head adjacency matrixes with ${T^l} \times {T^l}$ shape, ${W_t^n}$ represents trainable $1\times 1$ convolution corresponding to the $n$-th head.}

Different from the traditional local convolution processing temporal information in grided shape, the proposed temporal graph convolution extracts temporal dynamic with the designed temporal relation graph. Thus, the proposed model is able to capture both adjacent and non-adjacent temporal features in skeleton sequences without the interference of action-irrelevant features. Furthermore, the proposed multi-head temporal graph convolution fuses multiple temporal features from different temporal graphs, which can capture sufficient temporal features for action recognition.

\subsection{Model architecture}

In this section, we will elaborately describe the framework of the proposed Temporal Enhanced GCN (TE-GCN), which is designed to jointly learn the spatial and temporal structures of skeleton sequence for action recognition.

{\noindent {\bfseries Network architecture.}} The complete framework of the TE-GCN, shown in Fig.~\ref{fig:Architecture}, is constructed using ST-GCN as backbone. The ST-GCN backbone is composed of 1 input layer and 9 graph convolution layers. Each layer is the combination of regular spatial graph convolution and regular $1$D temporal convolution in a cascading manner. We replace regular $1$D temporal convolution with our proposed temporal enhanced graph convolution. The overall operation can be formulated as:
\begin{equation}\label{equation6}
{f_{out}^{st} = \sum\limits_n^N {{\rm{A}}_t^n\left[ {\sum\limits_k^{K_s} {{{\mathop{\rm W}\nolimits} _k}{f_{in}}(\Lambda _k^{ - \frac{1}{2}}{{{\rm{\tilde A}}}_{\rm{k}}}\Lambda _k^{ - \frac{1}{2}} \odot {M_k})} } \right]} {{\mathop{\rm W}\nolimits} ^n}},
\end{equation}
where $N$ denotes the number of heads of temporal graph. And $K_s$ represents the number of partition subsets in spatial graph. All the parameters have been elaborately described in the above section.
\begin{figure}
	\centering
	\includegraphics[width=1.01\linewidth]{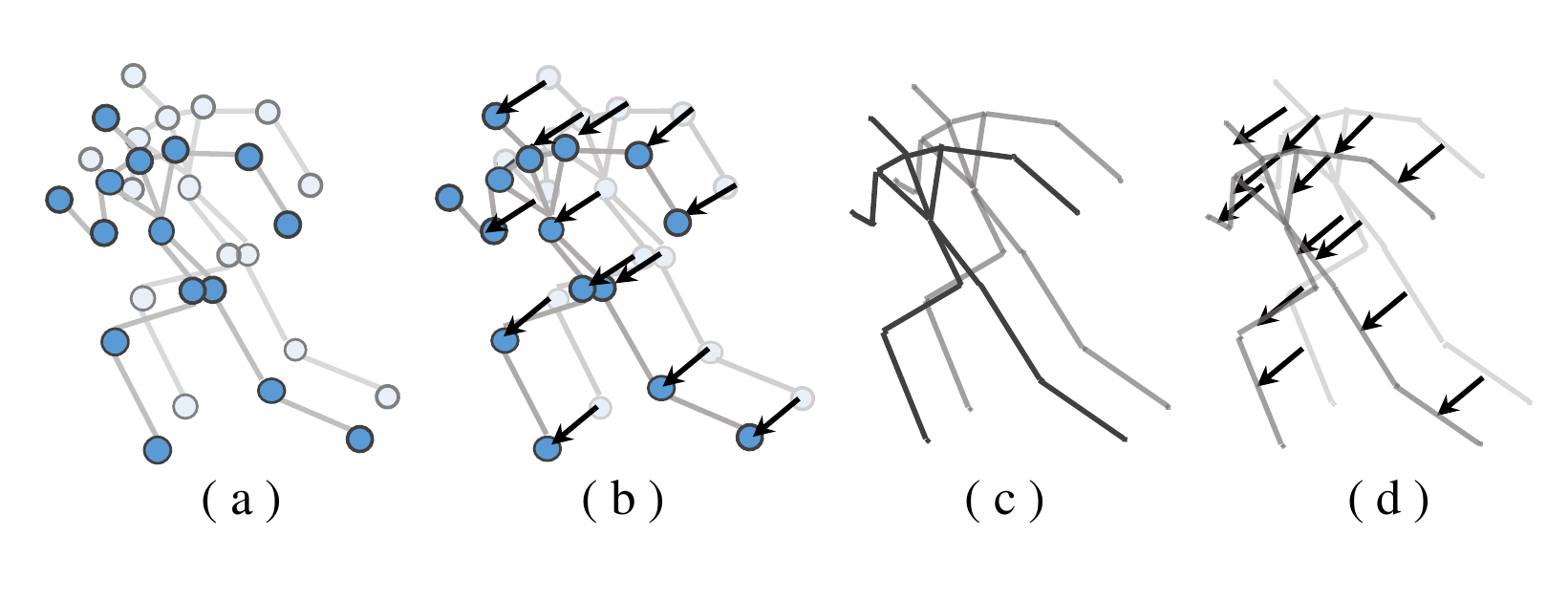}
	\caption{The illustration of different modalities that contain joint, bone and their corresponding motion modalities. (a) and (c) represent the joint and bone modalities. (b) and (d) represent their corresponding motion modalities. These modalities are complementary to each other for skeleton-based action recognition.}
	\label{fig:multi-modality}
\end{figure}
In contrast to the regular GCN that only considers the spatial structure information of skeleton graph in individual frames, the proposed TE-GCN simultaneously constructs the spatial and temporal graph for sufficient feature representation of action recognition. Besides, each layer of the TE-GCN is connected in residual form, which enables it to be inserted into any layer of existing models without destroying its initial behavior.
Global average pooling layer and a fully-connected layer along with softmax are appended after the temporal enhanced graph convolution layers for final classification.

It is worth noting that the proposed temporal enhanced graph convolution could be plugged into any layer of the network architectures for action recognition.
However, we have found it is best to insert it towards the end. This may be because that each temporal feature in bottom layer just represents several frames in receptive field, which is lack of semantic information.
Instead, the temporal features in top layer, with receptive field being enlarged, may be suitable to represent the temporal semantic instances.
Thus, the temporal graph convolution performed on top layer is able to capture temporal relation between temporal instances. We verify this phenomenon in the ablation study (Section~\ref{subsection53}).

{\noindent {\bfseries Multi-stream network with multi-modalities.}}  Although the skeleton joints contain important information about the action, the length and direction of bones as well as their corresponding motion modalities are also very important to provide some supplementary information. To further boost the performance, we explore these multiple modalities for action recognition inspired by the two-stream methods in \cite{shi2019two, shi2019skeleton}.  The proposed multi-stream network contains joint, bone and their corresponding motion modalities, which are illustrated in Fig.~\ref{fig:multi-modality}.
Specifically, the bone is represented as the vector pointing from the source joint to the corresponding target joint. The joint closer to the center of gravity of skeleton is defined as source joint and the joint farther away from the gravity center is target joint. Mathematically, given the source joint ${v_{1,t}} = ({x_{1,t}},{y_{1,t}},{z_{1,t}})$ and the target joint ${v_{2,t}} = ({x_{2,t}},{y_{2,t}},{z_{2,t}})$ in frame $t$, the bone is calculated as ${b_{{v_{1,t}},{v_{2,t}}}} = ({x_{2,t}} - {x_{1,t}},{y_{2,t}} - {y_{1,t}},{z_{2,t}} - {z_{1,t}})$.
As for motion information, it is obtained by calculating the movement information of the same joint or bone between the consecutive frames. Formally, the movement of joint $v$ in frame $t$ is calculated as ${M_{{v_t}}} = {v_{t + 1}} - {v_t}$.
The movement of bone is defined similarly as ${M_{{b_t}}} = {b_{t + 1}} - {b_t}$. These different modalities of skeleton data are fed into multi-stream network to make the prediction. For aggregation of multi-modalities,  the softmax scores of each modality are fused by weighted sum.

\section{Experiments}
To validate the effectiveness of the proposed TE-GCN network, extensive experiments are conducted on two skeleton-based action recognition datasets: NTU-60 RGB+D and NTU-120 RGB+D datasets. In the following, we first describe the datasets and implementation details in Section~\ref{subsection51} and Section~\ref{subsection52}, respectively.
Then, the exhaustive ablation study in Section~\ref{subsection53} are performed on NTU-60 RGB+D dataset to examine the effectiveness of the proposed components. Finally, our final model is evaluated and compared with the current state-of-the-art approaches in Section~\ref{subsection54}.

\subsection{Datasets} \label{subsection51}
{\noindent {\bfseries NTU-60 RGB+D:}}  The NTU-60 RGB+D~\cite{shahroudy2016ntu} is the most widely used dataset for action recognition. It contains 56,880 video samples and consists of 60 action classes performed by 40 subjects. There are three Microsoft Kinect v.2 cameras for each action from different horizontal angles: $-45^{\circ}$,
$0^{\circ}$, $-45^{\circ}$. This dataset provides 4 different modalities of data: RGB videos, depth map sequences, 3D skeleton data and infrared videos. Here, we focus on the 3D skeleton data. This skeleton data provides 3D coordinates of 25 joints for each person, and no more than 2 subjects in each clip. We follow two standard protocols, recommended by the original paper~\cite{shahroudy2016ntu}, for evaluation: (1) Cross-Subject (CS): half of the subjects in this dataset are used for training(40,320 samples) and remaining subjects are reserved for testing (16,560 samples). (2) Cross-View (CV): all samples of camera view 2 and 3 are chosen as the training set (37,920 samples) and all samples of camera view 1 are chosen as the testing set (18,960 samples). The Top-1 accuracy is reported on both benchmarks.

{\noindent {\bfseries NTU-120 RGB+D:}} The NTU-120 RGB+D~\cite{Liu_2019_NTURGBD120} is currently the largest indoor dataset with 3D joint coordinates for action recognition, which is the extended version of the NTU-60 RGB+D. It contains 114,480 action samples over 120 action classes, performed by 106 distinct subjects. This dataset contains 32 setups, and each setup has a specific location and background. The author of the original paper~\cite{Liu_2019_NTURGBD120} recommends two standard protocols for evaluation: (1) Cross-Subject (C-subject): training data comes from 53 subjects (63,026 samples), and the remaining 53 subjects are used for testing (50,922 samples). (2) Cross-Setup (C-setup): picking all the samples with even setup IDs for training (54,471 videos), and the remaining samples with odd setup IDs for testing (59,477 videos).

\subsection{Implementation details}\label{subsection52}
All the experiments are conducted on the PyTorch deep learning platform~\cite{paszke2017automatic}. Every sequence is transformed into a fixed length sequence by padding as input. We set the fixed length T as 300 for both NTU-60 RGB+D and NTU-120 RGB+D datasets.
Mini-batch stochastic gradient descent (SGD) optimizer~\cite{bottou2010large} is adopted to train the network and the initial learning rate here is set as 0.1 which will reduce by a factor 0.1 after epoch 40, 80 and 120. The batch size is set to 64. The weight decay of NTU-60 RGB+D and NTU-120 RGB+D dataset is set to 0.0005 and 0.001, respectively.
The model is trained for 150 epochs in total.

\subsection{Ablation study}\label{subsection53}
We conduct ablation study on the NTU-60 RGB+D dataset to verify the effectiveness of the components in the proposed TE-GCN.

{\noindent {\bfseries (1) Data preprocessing.}}

As described above, NTU-60 RGB+D dataset provides no more than 2 subjects in each clip. However, the bodies of skeleton data captured by Kinect cameras are prone to be more than 2, some of which are caused by the interference of the background. To filter out the incorrect skeleton, we first delete the extra-detected skeleton data according to the motion information. The motion information of skeleton body is defined as the sum of the variances of all joints across each of coordinate dimensions.
Specifically, if the motion value exceeds the threshold range, the skeleton body will be discarded. During the preprocessing, the threshold range is set to $\left[ {0.1,2.0} \right]$.
Then, the skeleton frame without skeleton body will be detected.
Finally, we translate the camera coordinate system to the body center of the first frame. It makes the network insensitive to the initial position of an action and normalize the skeleton data.
In detail, the skeleton data is represented in the camera coordinate system, whose coordinate origin is located at the position of the camera sensor. we subtract the coordinates of the ``spine joint'' in the first frame from the coordinates of each joint.
To verify the effectiveness of data processing, we compare the performance of ST-GCN using processed data with that using original data in Table~\ref{tab:dataprocess}. The results show that the data preprocessing considerably helps the recognition.

\begin{table}[H]
	\centering
	\renewcommand\arraystretch{1.8}
	\caption{Comparisons of the recognition accuracy using data processing on NTU-60 RGB+D skeleton dataset.}
	\label{tab:dataprocess}
	\begin{tabular}[\linewidth]{|c|c|c|}
		\hline
		\bfseries{Methods}	& \bfseries{CS (\%)} & \bfseries{CV (\%)} \\
		\hline
		Original	& $81.5$ & $88.3$ \\
		\hline
		Data preprocessing & $84.4$ & $92.1$ \\
		\hline
	\end{tabular}
\end{table}

{\noindent {\bfseries (2) Effectiveness of temporal graph convolution.}}

We first conduct experiments to verify the effectiveness of the proposed temporal graph convolution. As described in Section~\ref{subsection42}, the proposed temporal graph is constructed by two relevant functions: Feature Calculated and Feature Learned. Thus, we will focus on validating the effectiveness of these two different functions.
In this paper, we compare our proposed methods (shown as Feature Learned and Feature Calculated) with the regular ST-GCN in Table~\ref{tab:temporal graph}.
It shows that the temporal graphs designed by different relevance functions bring notable improvement for the action recognition task. Besides, the performance of Feature Calculated is slightly better. We speculate the reason is that the method of Feature Learned implicitly learns the relevance without any constrains. And compared with Feature Calculated, Feature Learned contains more learnable parameters, which make it difficult to learn the optimal correlations between temporal features.  In contrast, the method of Feature Calculated explicitly computes the correlations between temporal features. The explicitly calculated correlation may be more accurate for recognition than implicitly learned correlation. Therefore, the method of Feature Calculated could obtain more effective temporal relation for action recognition  with NTU-60 dataset.

\begin{table} [t]
	\centering
	\renewcommand\arraystretch{1.8}
	\caption{Ablation study on different relevance functions on NTU-60 RGB+D skeleton dataset under Cross-Subject protocol.}
	\label{tab:temporal graph}
	\begin{tabular}[\linewidth]{|c|c|}
		\hline
		\bfseries{Functions} & \bfseries{CS (\%)} \\
		\hline
		ST-GCN & $84.36$ \\
		\hline
		Feature Learned & $86.12$ \\
		\hline
		Feature Calculated & $86.43$ \\
		\hline
	\end{tabular}
\end{table}

{\noindent {\bfseries (3) Effectiveness of multi-head. }}

\begin{figure} [H]
	\centering
	\includegraphics[width=1\linewidth]{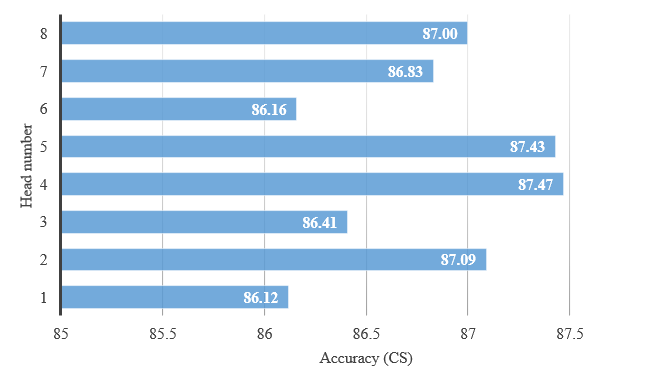}
	\caption{The performance of different head numbers of temporal graph on NTU-60 RGB+D skeleton dataset under Cross-Subject protocol.}
	\label{fig:multiHead}
\end{figure}

To justify the effectiveness of multi-head temporal graph, we examine the performance of the proposed method with different number of heads.
As shown in Fig.~\ref{fig:multiHead}, we can observe that the construction of multi-head temporal graph has further gains compared with only one head temporal graph. The intuition behind this is that multi-head temporal graph could capture multiple kinds of temporal relations. These multiple relationships will show different importance in learning discriminative temporal features, which facilitate the network learning more comprehensive temporal features.
Another observation is that continuously increasing the number of heads will saturate the recognition performance. It indicates that the unlimited accumulation of semantic information may cause confusion. Therefore, in the following experiments, we select to use 4 heads in proposed temporal enhanced graph convolution for more precise recognition.

{\noindent {\bfseries (4) Which layer to add temporal graph convolution?}}

Then, to study the effect of building temporal graph in different layers, we conduct experiments of plugging temporal graph convolution into different layers of the backbone network. Theoretically, the proposed temporal graph convolution can be inserted in any layer of the network.
As shown in Table~\ref{tab:layer}, we consider inserting temporal enhanced graph convolution into 9 different layers of ST-GCN from bottom-level to top-level. We could find that it is best to insert temporal graph convolution towards the top.
It might be because that the output feature maps of top layer contain more semantic information about temporal instances.
\begin{table*}
	\centering
	\renewcommand\arraystretch{1.8}
	\caption{Performance comparisons for building temporal graph in different layers on NTU-60 RGB+D with Cross-Subject protocol.}
	\label{tab:layer}
	\begin{tabular}[\linewidth]{|c|c|c|c|c|c|c|c|c|c|}
		\hline
		\multirow{2}{*}{\bfseries{Layer}} & \multicolumn{3}{c|}{\bfseries{Bottom}} & \multicolumn{3}{c|}{\bfseries{Mid}} & \multicolumn{3}{c|}{\bfseries{Top}}\\
		\cline{2-10}
		& $1$ & $2$ & $3$ & $4$ & $5$ & $6$ & $7$ & $8$ & $9$ \\
		\hline
		\bfseries{CS} & $85.6$ & $87.3$ & $87.3$ & $83.2$ & $86.8$ & $85.8$ & $86.0$ & $85.8$ & $87.4$ \\
		\hline
	\end{tabular}
\end{table*}

{\noindent {\bfseries (5) Contribution of multi-stream fusion.}}

Here, we examine the performance of our method using four different modalities in Table~\ref{tab:modalities}. Clearly, the multi-stream fusion outperforms the single-stream method. This demonstrates that our proposed method can be generalized to other input modalities. Specifically, by combining the joint and bone modalities, it brings improvement on NTU-60 RGB+D dataset under CV as well as CS protocols. Furthermore, the spatial and motion modalities added together still bring improvement. It suggests the complementarity of these four different modalities.

\begin{table} [H]
	\centering
	\renewcommand\arraystretch{1.7}
	\caption{Comparisons of multiple modalities, including joint, bone information and their corresponding motion modalities as well as fusion of all modalities.}
	\label{tab:modalities}
	\begin{tabular}[\linewidth]{|c|c|c|c|c|c|}
		\hline
		\multicolumn{2}{|c|}{\bfseries{Joint}} & \multicolumn{2}{c|}{\bfseries{Bone}} & \multirow{2}{*}{\bfseries{CS (\%)}}	& \multirow{2}{*}{\bfseries{CV (\%)}} \\
		\cline{1-4}
		\bfseries{spatial}	& \bfseries{motion} & \bfseries{spatial} & \bfseries{motion} &  &  \\
		\hline
		\checkmark & & & & $87.4$ & $93.4$ \\
		\hline
		& \checkmark & & & $87.3$ & $94.1$ \\	
		\hline
		&  & \checkmark & & $88.5$ & $93.3$ \\		
		\hline
		&  &  & \checkmark & $87.7$ & $93.1$ \\		
		\hline
		\checkmark & \checkmark &  &  & $88.9$ & $95.1$ \\				
		\hline
		& & \checkmark & \checkmark & $90.3$ & $95.2$ \\			
		\hline
		\checkmark & \checkmark & \checkmark & \checkmark & $90.84$ & $96.2$ \\		
		\hline
	\end{tabular}
\end{table}

\subsection{Comparison with the state-of-the-art methods}\label{subsection54}

We show the comparisons of our proposed final model against the existing deep-learning based methods on both NTU-60 RGB+D and NTU-120 RGB+D datasets for skeleton-based action recognition. The comparisons of recognition accuracy on both datasets are shown in Table~\ref{tab:NTU60 state-of-art} and Table~\ref{tab:NTU120 state-of-art}, respectively. The comparative methods include CNN-based, LSTM-based and GCN-based methods.

{\noindent\textbf{Results on NTU-60 RGB+D dataset.}
Compared with CNN-based~\cite{ke2018learning} and LSTM-based~\cite{zhang2017view} methods, our proposed TE-GCN surpasses by a large margin $9.7\%$ and $8.8\%$ on NTU-60 RGB+D with CS and CV protocols, respectively.
Since Yan et al.~\cite{yan2018spatial} first introduce GCN for skeleton-based action recognition, many GCN-based methods~\cite{tang2018deep, si2018skeleton,li2019spatio, li2019actional,peng2020learning,shi2020skeleton} have been proposed recently. These GCN-based methods achieve significant performance improvement than CNN-based and LSTM-based methods, which indicates the effectiveness of GCN in handling skeleton sequence for action recognition. Furthermore, our proposed multi-stream TE-GCN brings overall performance to $90.8\%$ and $96.2\%$ on NTU-60 RGB+D datasets with CS and CV protocols, respectively.}

\begin{table}
	\centering
	\renewcommand\arraystretch{1.8}
	\caption{Comparisons of the recognition accuracy (\%) with the state-of-the-art methods on NTU-60 RGB+D skeleton dataset.}
	\label{tab:NTU60 state-of-art}
	\begin{tabular}[\linewidth]{|c|c|c|c|}
		\hline
		\bfseries{Methods}	& \bfseries{CS} & \bfseries{CV} &	\bfseries{Conference}\\
		\hline
		VA-LSTM \cite{zhang2017view} & $79.2$ & $87.7$ &	CVPR2017\\
		\hline
		RotClip+MTCNN \cite{ke2018learning} & $81.1$ & $87.4$ & TIP2018\\
		\hline
		ST-GCN \cite{yan2018spatial} & $81.5$ & $88.3$ & AAAI2018\\
		\hline
		DPRL \cite{tang2018deep} & $83.5$ & $89.8$ & CVPR2018\\
		\hline
		SR-TSL \cite{si2018skeleton} & $84.8$ & $92.4$ & ECCV2018\\
		\hline
		STGR-GCN \cite{li2019spatio} & $86.9$ & $92.3$ & AAAI2019\\
		\hline
		AS-GCN \cite{li2019actional} & $86.8$ & $94.2$ & CVPR2019\\
		\hline
		NAS-GCN \cite{peng2020learning} & $89.4$ & $95.7$ & AAAI2020\\
		\hline
		MS-AAGCN \cite{shi2020skeleton} & $90.0$ & $96.2$ & TIP2020\\
		\hline
		\textbf{MS TE-GCN(our)} & $\textbf{90.8}$ & $\textbf{96.2}$ & -\\
		\hline
	\end{tabular}
\end{table}

{\noindent\textbf{Results on NTU-120 RGB+D dataset.}
On the currently largest indoor action recognition dataset, our model achieves $84.4\%$ for C-subject protocol and $85.9\%$ for C-setup protocol. Compared to CNN-based~\cite{ke2018learning} and LSTM-based~\cite{liu2016spatio} methods, our approach boosts the performance significantly. Compared to GCN-based methods~\cite{si2018skeleton, li2019actional,shi2019two, song2020richly}, our proposed TE-GCN achieves a more significant superiority.}

\begin{table}
	\centering
	\renewcommand\arraystretch{1.8}
	\caption{Comparisons of the recognition accuracy (\%) with the state-of-the-art methods on NTU-120 RGB+D skeleton dataset.}
	\label{tab:NTU120 state-of-art}
	\begin{tabular}[\linewidth]{|c|c|c|c|}
		\hline
		\bfseries{Methods}	& \bfseries{C-subject} & \bfseries{C-setup} &	\bfseries{Conference}\\
		\hline
		ST-LSTM \cite{liu2016spatio} & $55.7$ & $57.9$ & ECCV2016\\
		\hline
		RotClip+MTCNN \cite{ke2018learning} & $61.2$ & $63.3$ & TIP2018\\
		\hline
		ST-GCN \cite{yan2018spatial} & $70.7$ & $73.2$ & AAAI2018\\
		\hline
		SR-TSL \cite{si2018skeleton} & $74.1$ & $79.9$ & ECCV2018\\
		\hline
		AS-GCN \cite{li2019actional} & $77.7$ & $78.9$ & CVPR2019\\
		\hline
		2S-AGCN \cite{shi2019two} & $82.5$ & $84.2$ & CVPR2019\\
		\hline
		3S RA-GCN \cite{song2020richly} & $81.1$ & $82.7$ & CSVT2020\\
		\hline
		\ \textbf{MS TE-GCN(our)} & $\textbf{84.4}$ & $\textbf{85.9}$ & -\\
		\hline
	\end{tabular}
\end{table}

In general, the proposed multi-stream TE-GCN can achieve the state-of-the-art performance on both datasets, which suggests the superiority of our model.

\section{Conclusion}
How to model complex temporal dynamic of skeleton sequence is the key challenge for skeleton-based action recognition. In this paper, we propose a Temporal Enhanced Graph Convolutional Network (TE-GCN) to construct temporal relation graph to capture complex temporal dynamic. Firstly, temporal relation graph is constructed to explicitly build connections between semantically related temporal features to model temporal relations between both adjacent
and non-adjacent time steps. Then, to further explore the sufficient temporal dynamic, multi-head mechanism is designed to investigate multi-kinds of
high-order temporal relations. Finally, we also explore the joint, bone as well as their corresponding motion modalities and employ multi-stream TE-GCN to further improve the performance. Experimental results demonstrate that the proposed final model achieves the state-of-the-art recognition performance. The graph modeling in temporal dimension may provide insights for future research on action recognition.

\end{document}